\documentclass[10pt,twocolumn,letterpaper]{article}

\usepackage{iccv}
\usepackage{times}
\usepackage{epsfig}
\usepackage{graphicx}
\usepackage{amsmath}
\usepackage{amssymb}
\usepackage{bm}
\usepackage{algorithm}
\usepackage{algorithmic}
\usepackage{graphicx} 
\usepackage{subfigure}
\usepackage{enumitem}
\usepackage{multirow}
\usepackage{subfigure}
\usepackage{booktabs}
\usepackage{threeparttable}

\usepackage[sort,nocompress]{cite}

\usepackage[breaklinks=true,bookmarks=false]{hyperref}

\iccvfinalcopy 

\def\p{{\mathbf p}}
\def\g{{\mathbf g}}
\def\V{{\mathbf V}}
\def\D{{\mathbf D}}
\def\th{{\boldsymbol{ \theta}}}
\def\alo{{\boldsymbol{ \alpha_o}}}

\ificcvfinal\fi

\begin{document}

\title{Direct Differentiable Augmentation Search}

\author{\textbf{Aoming Liu\textsuperscript{\rm 1}, Zehao Huang\textsuperscript{\rm 2}, Zhiwu Huang\textsuperscript{\rm 1}, Naiyan Wang\textsuperscript{\rm 2}}\\
\textsuperscript{\rm 1}ETH Z\"urich, Switzerland,
\textsuperscript{\rm 2}TuSimple, Beijing\\
\{aoliu@student, zhiwu.huang@vision.ee\}.ethz.ch, \{zehaohuang18,winsty\}@gmail.com
}

\maketitle
\ificcvfinal\thispagestyle{empty}\fi

\newcommand{\ddaa}{DDAS}
\newcommand{\ddaaFull}{Direct Differentiable Augmentation Search}
\newcommand{\policy}{\varphi_{l}}
\newcommand{\lval}{\mathcal{L}_{val}}
\newcommand{\ltrain}{\mathcal{L}_{train}}
\newcommand{\MBE}{\mathbb{E}}
\newcommand{\MG}{\mathcal{G}}
\newcommand{\hatg}{\hat{g}}
\newcommand{\Mg}{\mathbf{g}}
\newcommand{\Mhatg}{\mathbf{\hat{g}}}

\begin{abstract}
Data augmentation has been an indispensable tool to improve the performance of deep neural networks, however the augmentation can hardly transfer among different tasks and datasets. Consequently, a recent trend is to adopt AutoML technique to learn proper augmentation policy without extensive hand-crafted tuning.
In this paper, we propose an efficient differentiable search algorithm called {\ddaaFull} (\ddaa). It exploits meta-learning with one-step gradient update and continuous relaxation to the expected training loss for efficient search. Our {\ddaa} can achieve efficient augmentation search without relying on approximations such as Gumbel-Softmax or second order gradient approximation.
To further reduce the adverse effect of improper augmentations, we organize the search space into a two level hierarchy, in which we first decide whether to apply augmentation, and then determine the specific augmentation policy. On standard image classification benchmarks, our {\ddaa} achieves state-of-the-art performance and efficiency tradeoff while reducing the search cost dramatically, e.g. 0.15 GPU hours for CIFAR-10. In addition, we also use {\ddaa} to search augmentation for object detection task and achieve comparable performance with AutoAugment\cite{cubuk2019autoaugment}, while being $1000 \times$ faster. Code will be released in \url{https://github.com/zxcvfd13502/DDAS_code}
\end{abstract}

\section{Introduction}
Due to the ``data hungry" nature of deep neural networks (DNN), data augmentation techniques, such as flipping, rotation, cropping, and color jittering, are essential tools to improve the performance. Data augmentation creates rich variation of data samples to reduce over-fitting issues caused by the high complexity of DNN. 
Although various hand-crafted data augmentation techniques~\cite{devries2017improved,zhang2018mixup,yun2019cutmix,inoue2018data,Xie_2020_CVPR,zhou2020data} are proposed recently, it's non-trivial to combine and adapt them when confronting a new task or dataset. This procedure usually requires expertise and extensive experiments to determine the optimal configuration. For example, an improper augmentation may not only be useless, but also may introduce harmful outliers in training.

AutoAugment (AA) \cite{cubuk2019autoaugment} is the pioneering work for automatic augmentation policy search. It utilizes an RL-based algorithm to search for optimal policy within a search space of 16 augmentation operations. During search, AA maximizes the accuracy on the validation set by optimizing 3 parameters: augmentation operation, its probability and magnitude, $(\mathtt{op}, \mathtt{prob}, \mathtt{mag})$. Despite its impressive performance on image classification and detection tasks, the search cost of AA is still prohibitive for democratizing the technique, e.g. it requires thousands of GPU hours just for searching on a small dataset like CIFAR-10. A popular approach is to make the optimization differentiable, which makes the gradient estimation more efficient. Following works, such as \cite{lim2019fast,ho2019population,li2020dada,hataya2019faster,chen2020hypernetwork}, manage to decrease the search cost to $0.1 - 0.2$ hours, but with visible performance degradation. 

\begin{figure}
	\centering
	\subfigure[CIFAR-100]{\includegraphics[width=0.45\linewidth]{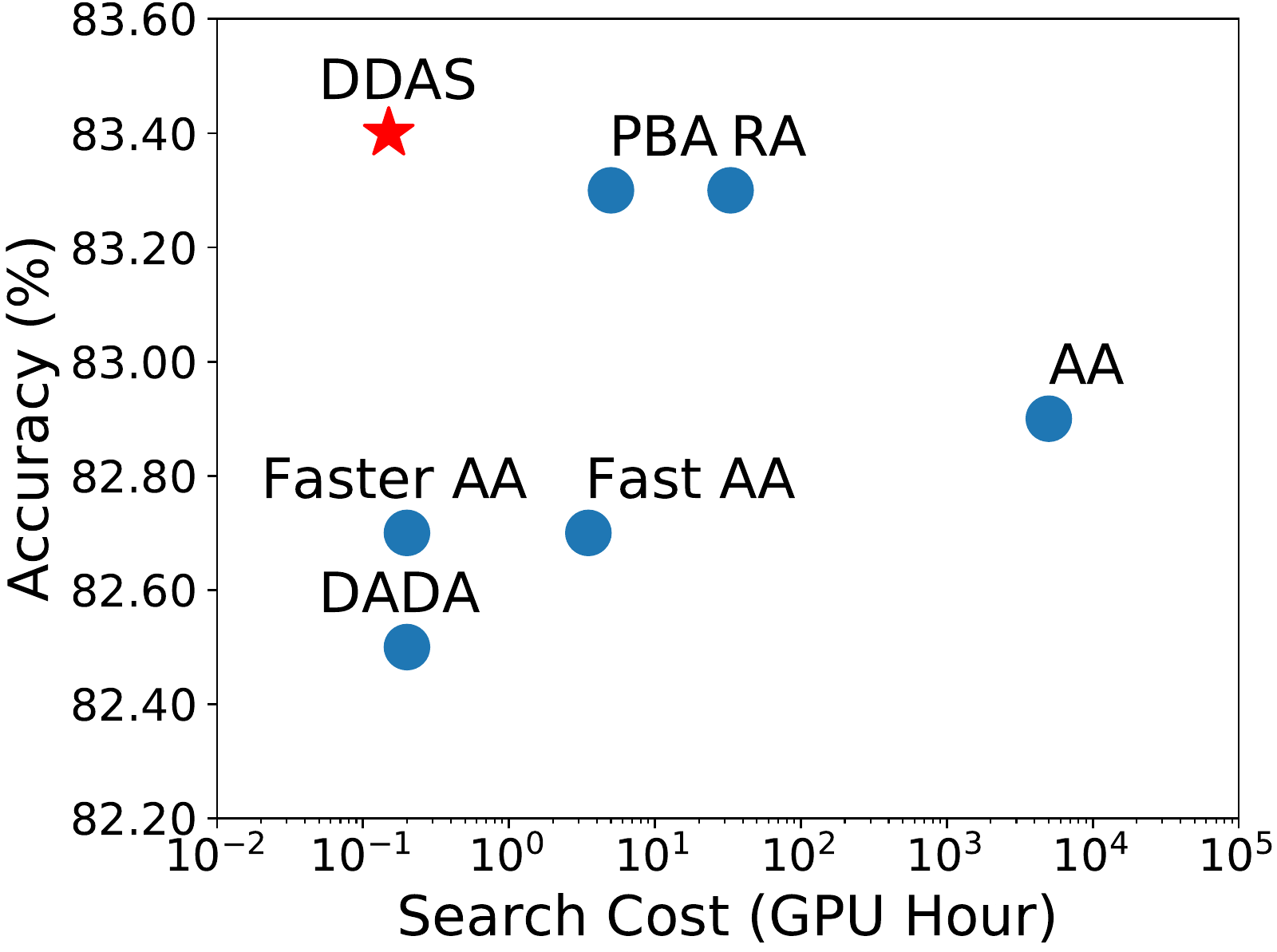}}
	\subfigure[ImageNet]{\includegraphics[width=0.49\linewidth]{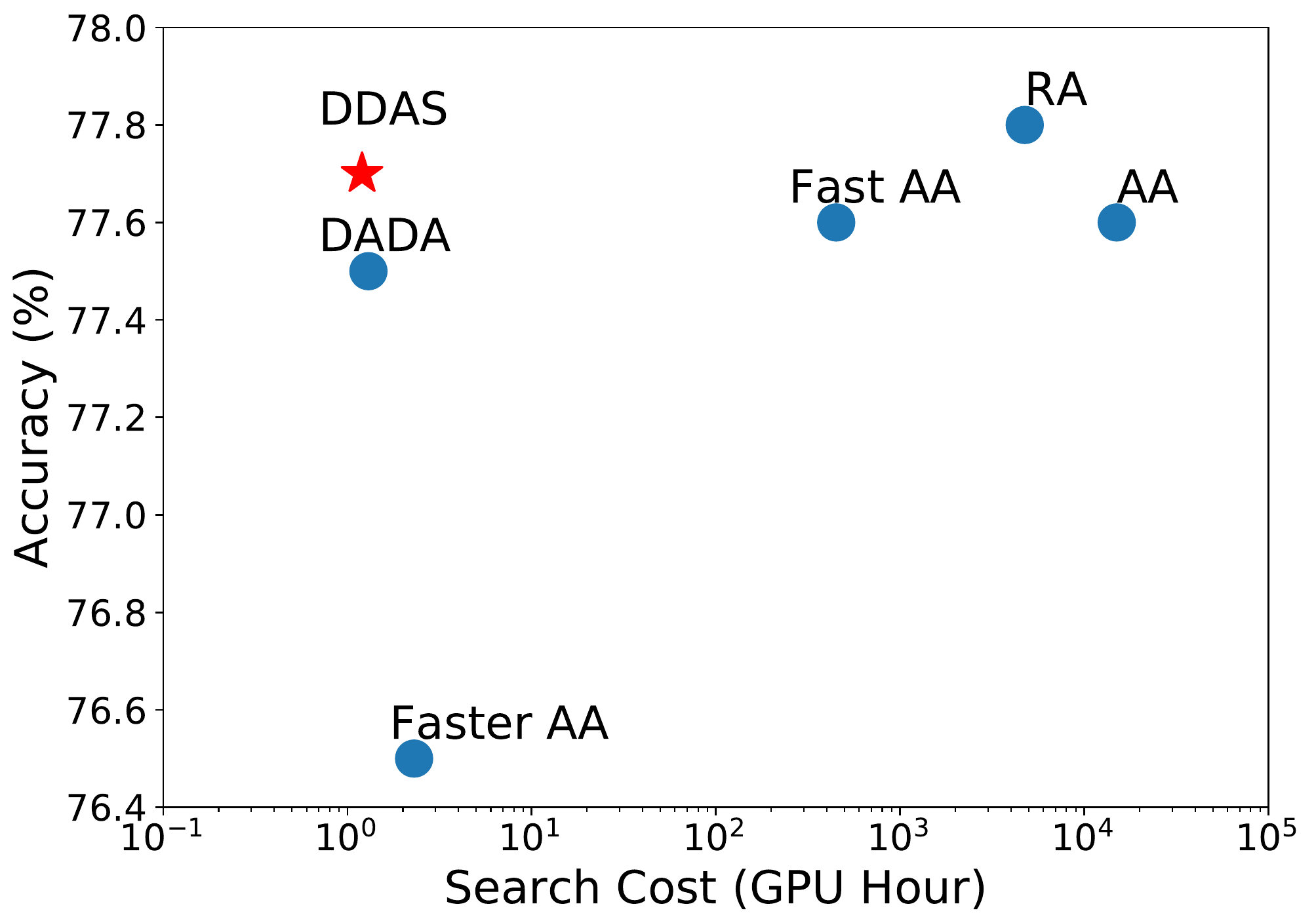}}
	\caption{Comparison between \ddaa\ and other state-of-the-art Automatic Augmentation works, including AA \cite{cubuk2019autoaugment}, RA\cite{cubuk2020randaugment}, PBA \cite{ho2019population}, Fast AA \cite{lim2019fast}, Faster AA \cite{hataya2019faster} and DADA \cite{li2020dada}. Higher accuracy and lower search cost (upper left) are preferred.  Our \ddaa\ is significantly more efficient while achieving even better performance.}
	\label{fig:cost-accuracy}
	\vspace{-10pt}
\end{figure}

On the other hand, AA only searches for a fixed augmentation policy while several works \cite{tian2020improving,he2019data} point out that dynamic augmentation policy may result in better performance. Following works such as OHL-Auto-Aug (OHLAA) \cite{lin2019online} or Adverisial Augment (AdvAA) \cite{zhang2020adversarial} propose online augmentation search manners that augmentation search is conducted together with training and different augmentation policies will be used for different training epochs. These online approaches achieve significant improvements over the offline counterpart, however their costs remain high, and the policies can hardly transfer to different training tasks.



To achieve faster search with better performance, we introduce \ddaaFull\ (\ddaa). Inspired by recent work on meta-learning\cite{liu2018darts,ren2018learning}, our goal is to find an augmentation policy which maximizes the network performance after one step gradient update. Inspired by OHLAA, we organize the augmentation policy into a two level hierarchy: we firstly determine the probability of augmenting the data, then we decide the probability of each augmentation operation. In this way, the policy has a chance to discard all augmentations. Then the differentiable search can be derived without tricks such as Gumbel-Softmax \cite{jang2016categorical} or second order gradient approximation \cite{liu2018darts}, as the probabilities naturally become weights of samples when considering the expectation of loss. 

We verify the effectiveness of our \ddaa\ on various models and datasets, including CIFAR-10/100 \cite{krizhevsky2009learning} and ImageNet \cite{deng2009imagenet}. All results show that our \ddaa\ can achieve competitive or better performance while dramatically reducing the search cost. In Fig.\ref{fig:cost-accuracy}, we visualize the comparison between \ddaa\ and other methods on CIFAR-100 and ImageNet. We can see that our proposed \ddaa~is on the Pareto optimal curve of the search cost v.s. testing accuracy. In addition, we also try to search for augmentation policy for object detection task. As far as we know, this task is rarely tackled due to its prohibitive cost. Thanks to our highly efficient search method, we could achieve results comparable with previous work \cite{zoph2019learning} that adopted AutoAugment for object detection, while costing $1000 \times$ less GPU hours. Our contributions can be summarized as follows:
 \begin{enumerate}
 \item We propose \ddaaFull\ (\ddaa), an efficient differentiable augmentation policy search algorithm. Through meta-learning with one-step gradient update, we can achieve efficient and effective augmentation search.
 \item We propose a compact yet flexible search space by explicitly modeling the probability of adopting augmentation. This design along with the epoch-wise policy reduces the adverse risk of aggressive augmentation.
 \item Besides the thorough evaluation experiments for image classification, we are the first work to demonstrate efficient augmentation search for object detection (20 GPU hours) is feasible.
 \end{enumerate}

\section{Related Works}
\subsection{Data Augmentation}
Data Augmentation has been a standard technique for deep neural network training. Common data augmentations include rotation, translation, cropping and resizing. In recent years, several novel augmentation operations are designed manually according to domain knowledge or intuitions, such as Cutout \cite{devries2017improved}, MixUp \cite{zhang2018mixup} and CutMix \cite{yun2019cutmix}. In addition to supervised learning, Data Augmentation is also important for semi-supervised learning \cite{Xie2019UnsupervisedDA,berthelot2019mixmatch,berthelot2019remixmatch}, self-supervised learning \cite{chen2020simple}, unsupervised learning \cite{zhao2020differentiable} and reinforcement learning \cite{kostrikov2020image}. Although these augmentations are effective on a specific task, transferring them into other tasks or datasets still requires extensive workload. Furthermore, improper augmentation may hurt the model performance. For example, \cite{cubuk2019autoaugment} reports that Cutout significantly hurts the  performance on reduced SVHN. Thus it is valuable to automatically search for suitable augmentation policies for different tasks.
\subsection{Automatic Data Augmentation}
Inspired by Neural Architecture Search (NAS) \cite{zoph2016neural,liu2018darts,xie2018snas}, recent works attempt to search for data augmentation policy automatically for different tasks, such as image classification \cite{cubuk2019autoaugment,ho2019population,lim2019fast,tang2020onlineaugment,wu2020generalization,hataya2020meta,tian2020improving,mounsaveng2020learning}, 2D object detection \cite{zoph2019learning,cubuk2020randaugment} and 3D point clouds related tasks \cite{cheng2020improving, li2020pointaugment}. AutoAugment (AA) \cite{cubuk2019autoaugment}, the pioneering work for automatic augmentation, proposes a novel search space consisting of 16 augmentation operations and adopts a reinforcement learning based algorithm for search. AA achieves promising performance improvement on classification task but the search cost is prohibitive (5000 GPU hours on CIFAR-10). Following AA, Population Based Augmentation (PBA) \cite{ho2019population} applies Population Based Training (PBT), an evolution based hyper-parameter optimization algorithm for augmentation policy search. Fast AutoAugment (Fast AA) \cite{lim2019fast} treats augmentation search as a density matching problem and uses Bayesian Optimization to solve it. Both of these two methods reduce the search cost from thousands of GPU hours to several hours, e.g. Fast AA only costs 3.5 GPU hours on CIFAR-10. 

More recently, Faster AutoAugment (Faster AA) \cite{hataya2019faster} and Differentiable Automatic Data Augmentation (DADA) \cite{li2020dada} propose differentiable relaxations to this challenging problem. DADA relaxes the original optimization with RELAX gradient estimator. Faster AA attempts to address this problem from another perspective: it replaced the original augmentation search task with a distribution matching task between the augmented data and original data. 
They both significantly improve the efficiency (0.1 - 0.2 GPU hour on CIFAR-10), but suffer from performance degradation compared with AA.

All the above methods need to firstly search for augmentation policy on an offline proxy dataset, and then transfer it to the final training. On the other hand, RandAugment (RA) \cite{cubuk2020randaugment}, OHL-Auto-Aug (OHLAA) \cite{lin2019online} and Adversarial AutoAugment (AdvAA) \cite{zhang2020adversarial} propose a different search paradigm. They argue that the augmentation policy searched on proxy task is sub-optimal for target task, and a fixed augmentation policy is suboptimal for different training stage. Specifically, RA directly uses naive grid search to find the best policy, while OHLAA and AdvAA propose online search manners by jointly optimizing augmentation policies and training the target networks. Recently, MetaAugment \cite{zhou2020metaaugment} further proposes sample-aware online data augmentation search. Despite their superior results,  their online search costs are quite large and not affordable on large dataset. 
Our proposed \ddaa\ is an offline method, however with the help of dynamic interpolation method, our method can apply different augmentation policies during different training stages. As a result, our method can get the best of two worlds.

\subsection{Comparison between DADA and \ddaa}
As mentioned above, DADA is close to our \ddaa. Thus we highlight some key differences here. Both DADA and \ddaa\ use meta-learning with one-step gradient update and differentiable optimization. However, as DADA follows the parameterization of AA, it has to use Gumbel-Softmax parameterization and RELAX gradient estimator \cite{grathwohl2017backpropagation} to make the optimization differentiable. In addition, second order gradient approximation \cite{liu2018darts} is necessary for DADA to achieve efficient search, which may result in inaccurate gradient approximation according to \cite{bi2020stabilizing}. In contrast, our \ddaa\ directly uses the expectation of training loss to derive the differentiable search formula without Gumbel-Softmax, gradient estimator and second order gradient approximation. 

A primary advantage of DDAS over DADA is the generalisability to complex tasks such as object detection. When applying DADA to object detection, the search cannot finally converge. We attribute this convergence problem to gradient noise and variance, which may be caused by inaccurate gradient approximation. In contrast, DDAS can achieve efficient augmentation search for object detection.


\section{Method}
\subsection{Search Space Reformulation}
As aforementioned, one notable difference between our method and previous works is that we explicitly model the probability of applying augmentation in our search space.
The search space of previous works indeed contains the option of no augmentation, however the augmentation is conducted in a multi-step manner. The probability of sampling each augmentation is independent at each step, consequently it may rarely sample a policy that applying no augmentation at all, and result in "over-strong" augmentation that deteriorates accuracy. 

\begin{figure}
\centering
\vspace{-0.3em}
\includegraphics[width=0.83\linewidth]{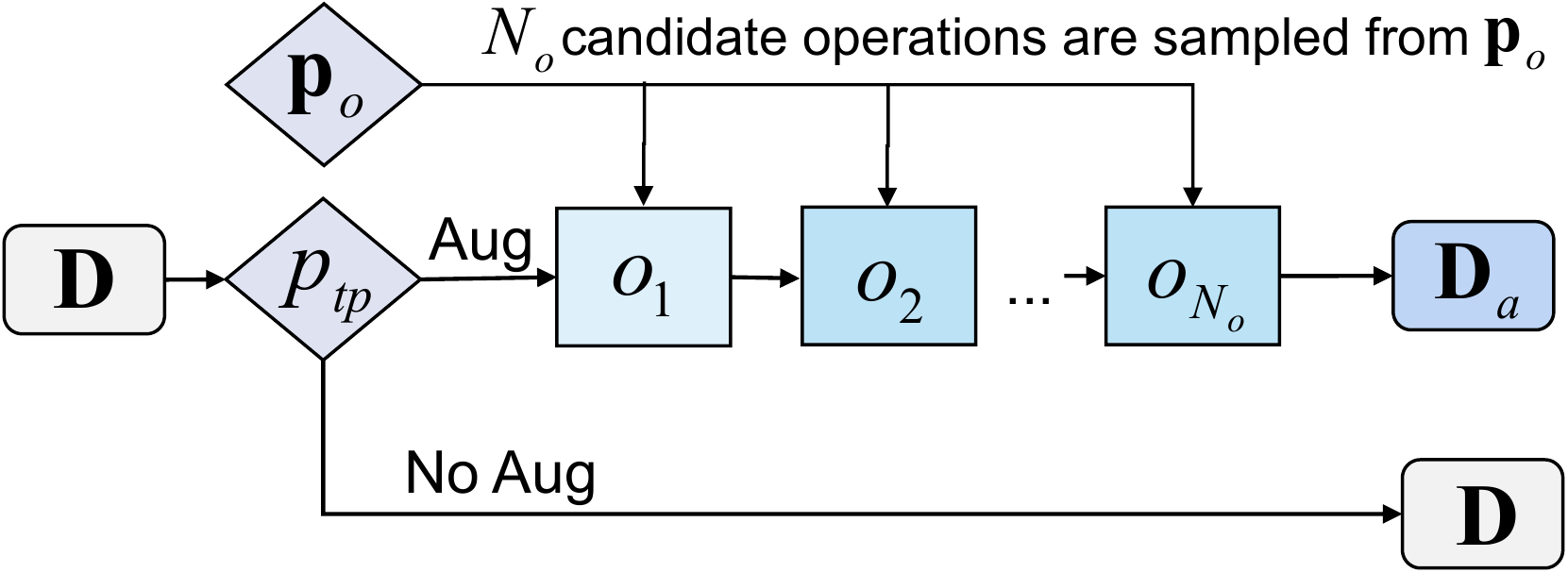}
\vspace{-0.3em}
\caption{Augmentation pipeline. First step is to decide whether to augment by sampling from $p_{tp}$. Second step is to decide $\mathtt{op}$ and $\mathtt{mag}$ by sampling from $\p_{o}$. In the second step, we sample $N_o$ operations from $K$ candidate operations and apply them sequentially for the input data.}
\label{fig:aug_pipeline}
\vspace{-1.6em}
\end{figure}

Formally, suppose we sample $N_o$ operations from $K$ candidate augmentation operations $\{o_1, ..., o_K\}$ sequentially to form an augmentation policy $\policy$. Thus the number of all possible augmentation policies is $L=K^{N_o}$.
Note that we discretize the continuous magnitude of augmentation and consider the same augmentation with different magnitudes as multiple operations. For example, $Rotation\ 15^\circ$ and $Rotation\ 20^\circ$ are regarded as two different candidate operations in our search space.
Specifically, we model the augmentation policy with two cascaded probabilities: \textbf{total probability} $p_{tp}$ and \textbf{operation probability} $\p_{o} = \{p_o^{o_1}, p_o^{o_2}, \dots, p_o^{o_K}\}$. $p_{tp}$ decides whether to apply augmentation on the original data and then $\p_{o}$ decides the probability for a certain augmentation.  
Fig.\ref{fig:aug_pipeline} shows the chain of constructing an augmentation policy.

\subsection{Formulation for Augmentation Search}
Augmentation search can be formulated as a bi-level optimization problem:
\begin{align}\label{object}
& \min_{\p} \mathcal{J}(\p) = \lval(\V, \th^*_{\p})
\\
{\rm{s.t.}} & \quad {\th^*_{\p}}  = \mathop{\arg}\mathop{\min} \limits_{\th} \MBE_{\p} \left[   \ltrain(\D, \th) \right], \nonumber
\end{align}
where $\ltrain$ and $\lval$ indicate the total training and validation loss; $\th$ represents the weights of DNN, and $\D$ and $\V$ denote the training samples and validation samples, respectively. Note that we assume that $\D$ in training is further augmented based on original data $\D_0$ and augmentation probability $\p = \{\p_o, p_{tp}\}$, while $\V$ for validation is kept as original as common practice.

It's non-trivial to solve this optimization problem since the inner optimization needs a full training of a given DNN, which is very time-consuming.
Inspired by DARTS\cite{liu2018darts}, we adopt the idea of meta-learning to relax Eqn.\ref{object} as a differentiable optimization problem. The inner optimization is approximated with one-step gradient update, instead of training until convergence. Thus we could get the approximated gradient of validation loss wrt. augmentation parameters:
\begin{equation}
\label{eq:aprox-val-grad}
\begin{aligned}
 &\nabla_{\p} \lval(\V, \th^{*}_{\p}) \\
 \approx & \nabla_{\p} \lval(\V,\th-\eta \cdot \nabla_{\th} \MBE_{\p}\left[\ltrain(\D, \th) \right]),
\end{aligned}
\end{equation}
where $\eta$ is the learning rate. 

Given an input mini-batch data $\D_0^t$ at each step $t$, we could generate a set of different augmented mini-batches $\{\D_1^t,...,\D_L^t\}$ by applying the $L$ augmentation policies, $\varphi_1,...,\varphi_L$, to it. We further define an augmentation policy $\varphi_l = \{o_l^1, \dots, o_l^{N_o}\}$ with $o_l^{i} \in \{o_1,\dots,o_K\}$ denoting the augmentation index chosen in the $i$-th step.
Then we can define the probability to get an augmented mini-batch $\D_l^t$ with augmentation policy $\varphi_l = \{o_l^1, \dots, o_l^{N_o}\}$ as 
\begin{align}
\label{eq:augbatchprob}
P(\D_l^t) = p_{tp} \cdot P(\varphi_l) = p_{tp} \cdot \prod_{i=1}^{N_o} p_o^{o_l^i},
\end{align}

Obviously, the probability of sampling a mini-batch without any augmentation can be represented as $P(\D_0^t)=1-\sum_{l=1}^{L} P(\D_l^t) = 1 - p_{tp}$.
Given the probability of each augmented mini-batch, we can expand the expected training loss as:
\begin{equation}
\begin{aligned}
\label{eq:loss-expect}
&\MBE_{\p} \left[\ltrain(\D_0^t, \th_{t})\right] \\
= &\sum_{l=1}^{L} P(\D_l^t) \ltrain(\D_l^t,\th^t) + P(\D_0^t)\ltrain(\D_0^t,\th^t).
\end{aligned}
\end{equation}

Then the gradient wrt. $\th^t$ can be easily derived by 
\begin{align}
\Mg^t &= \sum_{l=1}^{L} P(\D_l^t) \frac{\partial \ltrain(\D_l^t,\th^t)}{\partial \th^t} + P(\D_0^t)\frac{\partial \ltrain(\D_0^t,\th^t)}{\partial \th^t}\nonumber\\
&= \sum_{l=1}^{L} P(\D_l^t) \Mg^t_l + (1-\sum_{l=1}^{L} P(\D_l^t)) \Mg^t_0.
\end{align}

\subsection{Meta-Learning with One-Step Gradient Update}

Following the approximation in Eqn.\ref{eq:aprox-val-grad}, network parameters are updated for one step with learning rate $\eta$:
\begin{align}
\label{eq:1-step-update}
\th^{t+1} = \th^{t} - \eta\cdot \Mg^t.
\end{align}

Then assume we sample a validation mini-batch $\V^t$ from validation dataset, the validation loss becomes $\lval(\V^t,\th^{t+1})$, and we can get the gradient of validation loss wrt. $\p_o$ and $p_{tp}$ as :
\begin{small}
\begin{align}
\label{eq:grad-pt}
\frac{\partial \lval(\V^t,\th^{t+1})}{\partial p_o^{o_k}} = \sum_{l=1}^{L} \frac{\partial \lval(\V^t,\th^{t+1})}{\partial P(\D_l^t)}\cdot \frac{\partial P(\D_l^t)}{\partial p_o^{o_k}}. 
\end{align}
\end{small}
\begin{small}
\begin{align}
\label{eq:grad-po}
\frac{\partial \lval(\V^t,\th^{t+1})}{\partial p_{tp}}=\sum_{l=1}^{L}  \frac{\partial \lval(\V^t,\th^{t+1})}{\partial P(\D_l^t)} \cdot \frac{\partial P(\D_l^t)}{\partial p_{tp}}.
\end{align}
\end{small}

The partial derivative of $\lval(\V^t,\th^{t+1})$ wrt. $P(\D_l^t)$ can be further derived by:
\begin{equation}
\begin{aligned}
\label{eq:original-grad}
\frac{\partial \lval(\V^t,\th^{t+1})}{\partial P(\D_l^t)} &=  \frac{\partial \lval(\V^t,\th^{t+1})}{\partial \th^{t+1}}^{\top} \cdot\frac{\partial \th^{t+1}}{\partial P(\D_l^t)} \\
 &=  \eta \cdot {\Mg^t_{val}}^{\top} \cdot (\Mg^t_0 - \Mg^t_l),
\end{aligned}
\end{equation}
where $\Mg^t_{val} = \frac{\partial \lval(\V^t,\th^{t+1})}{\partial \th^{t+1}}$.
Then Eqn.\ref{eq:grad-pt} and Eqn.\ref{eq:grad-po} can be reformulated as:
\begin{equation}
\label{eq:deri-grad-pt}
\frac{\partial \lval(\V^t,\th^{t+1})}{\partial p_o^{o_k}}= 
\sum_{l=1}^{L}\eta \cdot  {\Mg^t_{val}}^{\top} \cdot (\Mg^t_0 - \Mg^t_l) \cdot p_{tp} \cdot \frac{\partial P(\policy)}{\partial p_o^{o_k}},
\end{equation}
\begin{equation}
\label{eq:deri-grad-po}
\frac{\partial \lval(\V^t,\th^{t+1})}{\partial p_{tp}}=
 \sum_{l=1}^{L}\eta \cdot  {\g^t_{val}}^{\top} \cdot (\g^t_0 - \g^t_l) \cdot P(\policy).
\end{equation}

We denote $\frac{\partial \lval(\V^t,\th^{t+1})}{\partial p_o^{o_k}}$ as $\Mg_{p_o^{o_k}}$ and $\frac{\partial \lval(\V^t,\th^{t+1})}{\partial p_{tp}}$ as $\Mg_{p_{tp}}$, then
$p_o^{o_k}$ and $p_{tp}$ can be updated by gradient descent with $\Mg_{p_o^{o_k}}$ and  $\Mg_{p_{tp}}$,  respectively.

\subsection{Stochastic Policy Sampling}
\begin{figure*}
\centering
\vspace{-1em}
\includegraphics[width=0.6\linewidth]{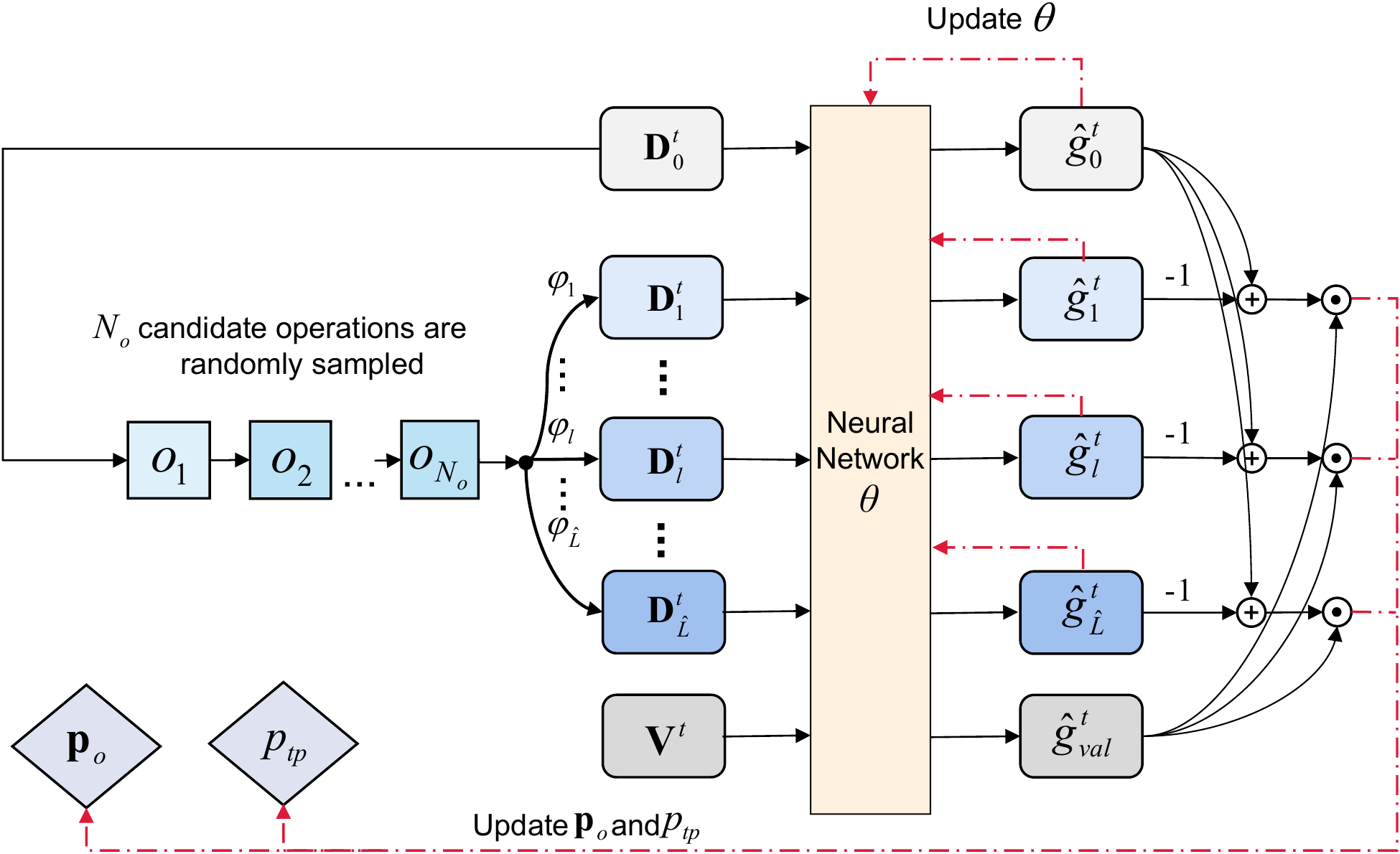}
\caption{Overview of our proposed Directly Differentiable Automatic Augmentation pipeline. $\hat{L}$ augmented mini-batches $\D_1^t,\dots,\D_{\hat{L}}^t$ are generated by applying augmentation policies $\varphi_{\hat{1}},...,\varphi_{\hat{L}}$ sampled with uniform probability distribution to original mini-batch $\D_0^t$. Augmentation parameters, $p_{tp}$ and $\p_o$, are updated with approximated gradients according to Eqn.\ref{eq:approx-grad-final-po} and Eqn.\ref{eq:approx-grad-final-pt}.}
\label{fig:rba_overview}
\vspace{-0.5em}
\end{figure*}

In practice, it will be intractable if we directly calculate the gradient of $\p$ following Eqn.\ref{eq:deri-grad-pt} and Eqn.\ref{eq:deri-grad-po}, as we need to sample all the $L$ possible augmentation polices where $L$ can be very large, e.g. $L = 4624$ for CIFAR experiments.
In order to search efficiently, we only randomly sample $\hat{L}$ policies $\{\varphi_{1},...,\varphi_{\hat{L}}\}$ to calculate the expected training loss as described in Eqn~\ref{eq:loss-expect}:
\begin{equation}
\begin{small}
\begin{aligned}
\label{eq:loss-expect-approx}
& \MBE_{\p} \left[\ltrain(\D_0^t,\th^t) \right] \\
\approx & \sum_{l=1}^{\hat{L}} \frac{P(\D_{l}^t)}{Z} \ltrain(\D_{l}^t,\th^t) + (1- \sum_{l=1}^{\hat{L}} \frac{P(\D_{l}^t)}{Z}) \ltrain(\D_0^t,\th^t),
\end{aligned}
\end{small}
\end{equation}
where $Z = \sum_{l=  1}^{\hat{L}}(P(\varphi_{l}))$ is a normalization factor to make sure that the probability of no augmentation equals to $1 - p_{tp}$. Based on Eqn.\ref{eq:loss-expect-approx}, we have: 
\begin{align}
\label{eq:approx-grad-1}
\Mg_{p_o^{o_k}} = & \frac{1}{Z} \cdot \sum_{l=1}^{{}\hat{L}}\eta \cdot  {{{}\hat{\Mg}}^t_{val}}^{\top} \cdot (\Mhatg^t_0 - \Mhatg^t_l)   \cdot p_{tp} \cdot \frac{\partial P(\policy)}{\partial p_o^{o_k}},   \\
\Mg_{p_{tp}} = & \frac{1}{Z} \cdot \sum_{{}\hat{l}=\hat{1}}^{{}\hat{L}}\eta \cdot  {{{}\hat{\Mg}}^{t}_{val}}^{\top} \cdot (\Mhatg^t_0 - \Mhatg^t_l) \cdot P(\policy),
\end{align}
where $\mathbf{\hat{g}}^t_{val}$ is the validation gradient obtained after using approximated loss expectation in Eqn.\ref{eq:loss-expect-approx} to do one-step gradient update, and $\Mhatg^t_l$ is the gradient of training loss corresponding to $\D_l^t$.
In the actual search process, we find that there is obvious scale difference on gradients caused by different data at each step, as each $\D_0^t$ could contribute differently to the validation loss. In order to achieve stable search, we normalize the gradients for $\D_0^t$ and $\{\D_1^t \dots \D_{\hat{L}}^t\}$ with a scale factor $Z_g$ calculated by different data at each step:
\begin{align}
\label{eq:g-norm}
Z_g = \sum_{l=1}^{\hat{L}} \left| {{{}\hat{\Mg}}^{t}_{val}}^{\top} \cdot (\Mhatg^t_0 - \Mhatg^t_l) \right|.
\end{align}
With this approximation, $\Mg_{p_o^{o_k}}$ and $\Mg_{p_{tp}}$ finally become:
\begin{small}
\begin{equation}
\label{eq:approx-grad-final-po}
\Mg_{p_o^{o_k}} =  \frac{1}{Z \cdot Z_g} \cdot \sum_{l=1}^{\hat{L}}\eta \cdot  {{{}\hat{\Mg}}^t_{val}}^{\top} \cdot (\Mhatg^t_0 - \Mhatg^t_l)   \cdot p_{tp} \cdot \frac{\partial P(\policy)}{\partial p_o^{o_k}},
\end{equation}
\begin{equation}
\label{eq:approx-grad-final-pt}
\Mg_{p_{tp}} =  \frac{1}{Z \cdot Z_g} \cdot \sum_{{}\hat{l}={}\hat{1}}^{{}\hat{L}}\eta \cdot  {{{}\hat{\Mg}}^{t}_{val}}^{\top} \cdot (\Mhatg^t_0 - \Mhatg^t_l) \cdot P(\policy).
\end{equation}
\end{small}

We find that we can achieve efficient optimization with the derived $\Mg_{p_o^{o_k}}$ and $\Mg_{p_{tp}}$, and there is no need to apply second order gradient approximation\cite{liu2018darts} to further improve the search efficiency.

\subsection{Implementation Details and Algorithms}
In this section, we will introduce implementation details of our proposed algorithm.

\subsubsection{Search Details} We search for augmentation policies on proxy tasks by splitting the original dataset into smaller train and validation dataset, $S_{train}$ and $S_{val}$. 
And the proxy search runs for $T_{max}$ epochs, which is smaller than that in common training. Note that we sample augmentations from a uniform distribution during search to avoid Matthew Effect: operations with large probabilities to be continuously selected and promoted.
\subsubsection{Probability Implementation} To make $p_{tp}$ fall within valid range, we implement it using sigmoid function with its input noted as $\alpha_{tp}$: $p_{tp}=\frac{1}{1+e^{-\alpha_{tp}}}$. To make elements of $\p_o$ sum to 1, we implement $\p_o$ using softmax function with its input noted as $\alo$: $p_o^{o_k}=\frac{e^{\alpha_o^k}}{\sum_{i=1}^Ke^{\alpha_o^i}}$, where $\alpha_o^k$ is the $k$-th item of $\alo$.

\subsubsection{Augmentation Operations and Magnitudes} Similar as \cite{lin2019online}, we combine $\mathtt{op}$ and $\mathtt{mag}$ by discrete sampling of magnitude for each operation. Each operation requiring magnitude is combined with the selected magnitude values to form independent candidate operations: $o_i = (\mathtt{op}_i,\ \mathtt{mag}_i)$. Other operations that do not need magnitude are regarded as single candidate operations $o_i = (\mathtt{op}_i, \mathtt{None})$. We follow the setting of $\mathtt{mag}$ in RandAugment \cite{cubuk2019autoaugment}, which is elaborated in the appendix detailedly.

%
%



\subsubsection{Dynamic Offline Policy} There are evidences suggesting that different training stage may prefer different augmentation policy \cite{he2019data,tian2020improving}. Consequently, we design a dynamic offline augmentation policy which saves the snapshot of augmentation parameters  $\p$ for every epoch during search, and then replay them during final re-training. However, the re-training typically takes more epochs than search, we further propose to upsample the policy with nearest neighbor interpolation. When applied to training stage, we use 1-D mean filter with size $F_s$ to smooth the $p_{tp}$ as there may be fluctuations of $p_{tp}$ which will deteriorate training accuracy. 

\subsubsection{\ddaa~ Overview and Algorithm} We summarize the whole method in Algorithm \ref{alg:search}, and a visual demonstration is shown in Fig. \ref{fig:rba_overview}.
\begin{algorithm}[ht]
	\caption{Search Algorithm of \ddaa}
	\small
	\begin{algorithmic}[1]
		\STATE INPUT: $S_{train}$ and $S_{val}$ from original dataset.
		\STATE Initialize $\p_o$, $p_{tp}$ and network parameter $\th$.
		\FOR{$T = 1,2,...,T_{max}$}
		\FOR{$t = 1,2,...,max\_iter$}
		\STATE Sample $\D_0^t$ from $S_{train}$ and compute $\Mhatg^t_0$.
		\STATE Sample $\V^t$ from $S_{val}$.
		\FOR{$l$ = 1,...,$\hat{L}$}
		\STATE Randomly sample $\varphi_{l}$ with equal probabilities.
		\STATE Apply $\varphi_{l}$ to $\D_0^t$ to generate $\D_{l}^t$.
		\STATE Compute $\Mhatg^t_l$ with the training loss of $\D_l^t$.
		\ENDFOR
		\STATE Do one-step gradient update and compute $\Mhatg^t_{val}$.
		\STATE Update $\p_o^*$ and $p_{tp}^*$ according to Eqn.\ref{eq:approx-grad-final-po} and Eqn.\ref{eq:approx-grad-final-pt}.
		\ENDFOR
		\STATE $\p_{o}(T) \xleftarrow{} \p_o^*$.  $p_{tp}(T) \xleftarrow{} p_{tp}^*$.
		\ENDFOR
		\STATE Smooth $p_{tp}(1),..,p_{tp}(T_{max})$ with mean filter.
		\RETURN $p_{tp}(1),..,p_{tp}(T_{max})$ and $\p_{o}(1),...,\p_{o}(T_{max})$.
	\end{algorithmic}
	\label{alg:search}
\end{algorithm}

\section{Experiments}

\begin{table*}
\small

\caption{
Comparison of search time in GPU hours.
}
\label{table:time}
\vspace{3pt}
\centering
\begin{threeparttable}  
\begin{tabular}{c|cccccc|c}
\toprule
         & AA    &RA$\dag$ & PBA & Fast AA &  Faster AA& DADA & \textbf{\ddaa} \\
\midrule
GPU & P100 &2080Ti& Titan XP & V100 & V100 & Titan XP & 2080Ti \\
\midrule
CIFAR-10/100 & $5000$  &$33$ & $5$   & $3.5$ &  $0.2$ & $0.1/0.2$ & $0.15$  \\
ImageNet & $15000$ &$4750$& -     & $450$  & $2.3$ & $1.3$ & $1.2$  \\
\bottomrule

\end{tabular}
\begin{tablenotes}
	\centering
	\item $\dag$ Estimated based on our experiment settings.
\end{tablenotes}
\end{threeparttable}  
\vspace{-7pt}
\end{table*}

We apply our \ddaa\ to search for augmentation for two tasks: image classification and object detection. First we perform a sanity check on a toy experiment to show that our \ddaa\ can learn reasonable  augmentation. Then, for image classification, we carry out experiments on three datasets: CIFAR-10, CIFAR-100 \cite{krizhevsky2009learning} and ImageNet \cite{deng2009imagenet}. For object detection, we evaluate our method on the popular COCO dataset \cite{lin2014microsoft}. 

We mainly compare our performance and search cost with offline augmentation search, including AA \cite{cubuk2019autoaugment}, Fast AA \cite{lim2019fast}, PBA \cite{ho2019population}, RA \cite{cubuk2020randaugment}, Faster AA \cite{hataya2019faster} and DADA \cite{li2020dada}. Note that AdvAA \cite{zhang2020adversarial}, OHLAA \cite{lin2019online} and MetaAugment \cite{zhou2020metaaugment} are not compared here as it is hard to compare with them in a fair setting. For example, online methods usually adopt large batch sizes, which has significant impact on the performance.
\subsection{Sanity Check}
\noindent \textbf{Experiment Setting}. To demonstrate the effectiveness and correctness of our {\ddaa}, we firstly design a toy example on the CIFAR-10 dataset. In normal training, it's hard to verify the correctness of an augmentation policy, as multiple augmentations may work in synergy to improve the performance. Therefore, we design the following sanity check: in contrast to the normal setting, we rotate the images in the validation set for $90^{\circ}$. And we add rotation for $90^{\circ}$ as a candidate augmentation, which is noted as \texttt{Rotate90}. Then obviously, a reasonable policy should assign higher probability to the \texttt{Rotate90} operation. For simplicity, we replace the original rotation operation with \texttt{Rotate90} and only sample one magnitude for each operation. Other details are listed in the appendix.

\begin{figure}
	\centering  
	\includegraphics[scale=0.35]{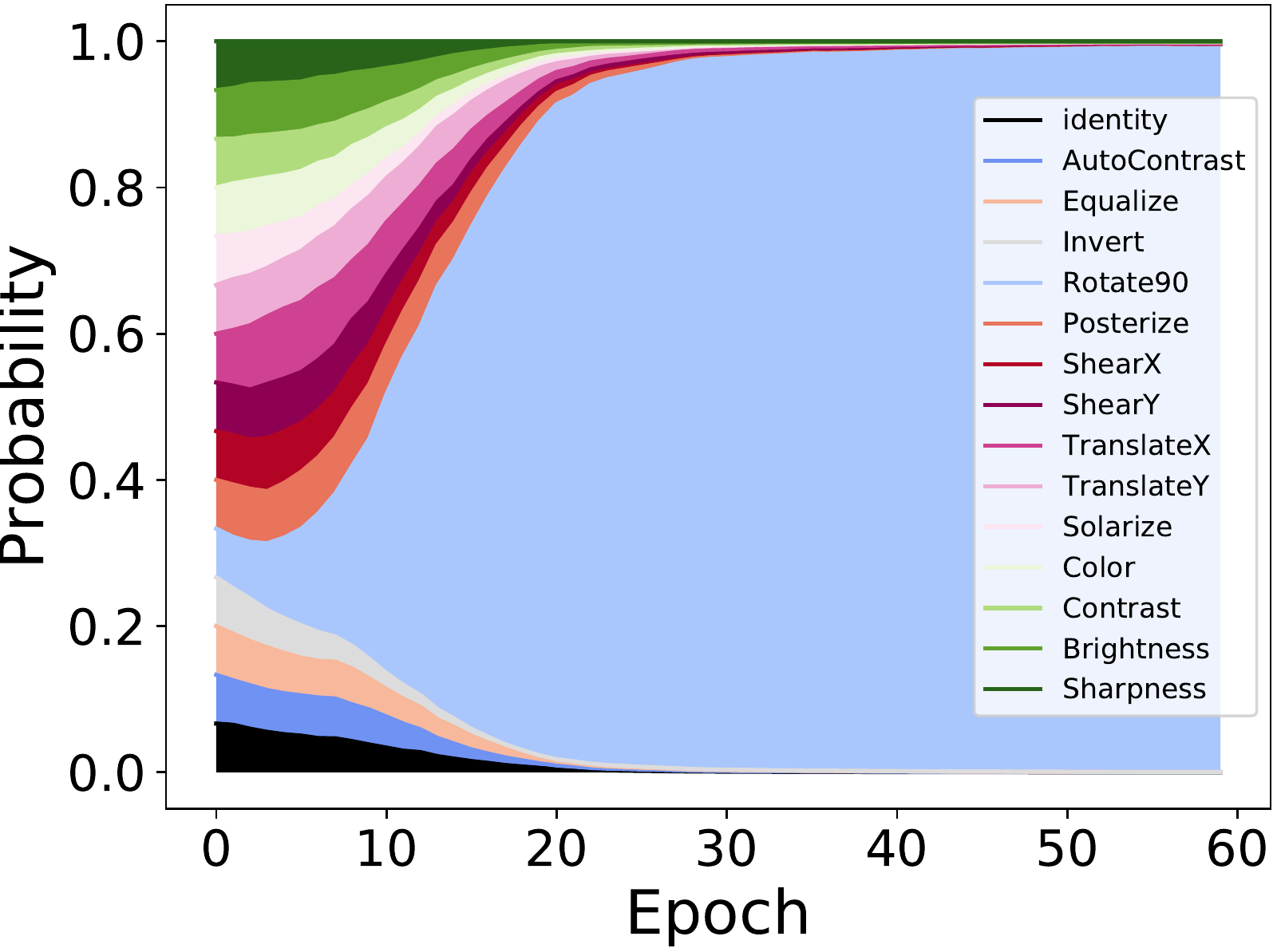}
	\caption{$\p_{o}$ searched in sanity check experiment. \texttt{Rotate90} is the dominant operation.}
	\label{fig:sc-po}
	\vspace{-10pt}
\end{figure}

\noindent \textbf{Result}. We visualize the $\p_o$ searched in Fig~\ref{fig:sc-po}. We can see that the \texttt{Rotate90} rapidly becomes the dominant operation, which proves the correctness of our \ddaa.

\subsection{CIFAR-10 and CIFAR-100}
\noindent \textbf{Search Setting}. For CIFAR-10 and CIFAR-100, we conduct experiments with two different networks. We firstly search for augmentation policy on a proxy task with a small network, Wide-ResNet-40-2\cite{zagoruyko2016wide}, on part of the dataset for 20 epochs, with 2000 images for training and 2000 images for validation. We run the search for 20 epochs on both CIFAR-10 and CIFAR-100. The training mini-batch size is 32 while the validation mini-batch size is 256. At each step, we sample $\hat{L}=3$ augmentation policies randomly according to a uniform distribution. There are $N_o = 2$ operations in an augmentation policy. The search is carried out on a RTX 2080Ti. We initialize $\p_o$ equally and $p_{tp}$ as 0.35.

\noindent \textbf{Training Setting}. After search, we apply the searched augmentation to two prevalent networks, Wide-ResNet-28-10\cite{zagoruyko2016wide} and Shake-Shake(26 2x96d)\cite{gastaldi2017shake}. We train the networks on the full training set and report the performance evaluated on test set. Wide-ResNet-28-10 is trained for 200 epochs and Shake-Shake(26 2x96d) is trained for 1800 epochs. 
We run every experiment for three times and report the average test error rate. Other details are in the appendix.

\begin{table*}
\small
\caption{
CIFAR-10 and CIFAR-100 test error rates $(\%)$. WRN and SS are the shorthand of Wide-ResNet and Shake-Shake respectively.
}
\vspace{3pt}
\label{table:c10_c100_svhn}
\centering
\begin{tabular}{l|cccccccc|c}
\toprule
 & Baseline & Cutout & AA & PBA & Fast AA & RA & Faster AA & DADA & \textbf{\ddaa} \\
\midrule
\textbf{CIFAR-10} & & & & & & & & & \\
WRN-28-10 & $3.9$ & $3.1$ & $2.6$ & $2.6$ & $2.7$ & $2.7$ & $2.6$ & $2.7$ & $2.7 \pm 0.1$ \\
SS {\scriptsize (26 2x96d)} & $2.9$ & $2.6$ & $2.0$ & $2.0$ & $2.0$ & $2.0$ & $2.0$ & $2.0$ & $2.1 \pm 0.1$ \\
\midrule
\textbf{CIFAR-100} & & & & & & & & & \\
WRN-28-10 & $18.8$ & $18.4$ & $17.1$ & $16.7$ & $17.3$ & $16.7$ & $17.3$ & $17.5$ & $16.6 \pm 0.2$ \\
SS {\scriptsize (26 2x96d)} & $17.1$ & $16.0$ & $14.3$ & $15.3$ & $14.9$ & - & $15.0$ & $15.3$ & $15.1 \pm 0.2$ \\
\bottomrule
\end{tabular}
\vspace{-10pt}
\end{table*}

\noindent \textbf{Result}. The search costs of {\ddaa} are listed in Table~\ref{table:time}. And the test error rates are shown in Table~\ref{table:c10_c100_svhn}. Results of other automatic augmentation search works are also listed for comparison. As shown in Table~\ref{table:time}, our {\ddaa} costs only 0.15 GPU hours to search on CIFAR-10 and CIFAR-100, which is significantly faster than AA, Fast AA and PBA. Comparing to those efficient search methods, such as Faster AA and DADA, our {\ddaa} yields better performances as shown in Table~\ref{table:c10_c100_svhn}. The advantage is even more significant in complicated datasets such as CIFAR-100. Jointly considering these two aspects, our method achieves the sweet spot for accuracy and efficiency tradeoff for automatic augmentation policy search.

Another desired property of our {\ddaa} is that it enables the transfer of augmentation policy searched with small network to large networks. Our experiments further verify this claim: The policy obtained using Wide-ResNet-40-2 can transfer well to Wide-ResNet-28-10 and Shake-Shake(26 2x96d).

\begin{figure}
	\centering
	\subfigure[CIFAR-10]{\includegraphics[width=0.48\linewidth, height=0.35\linewidth]{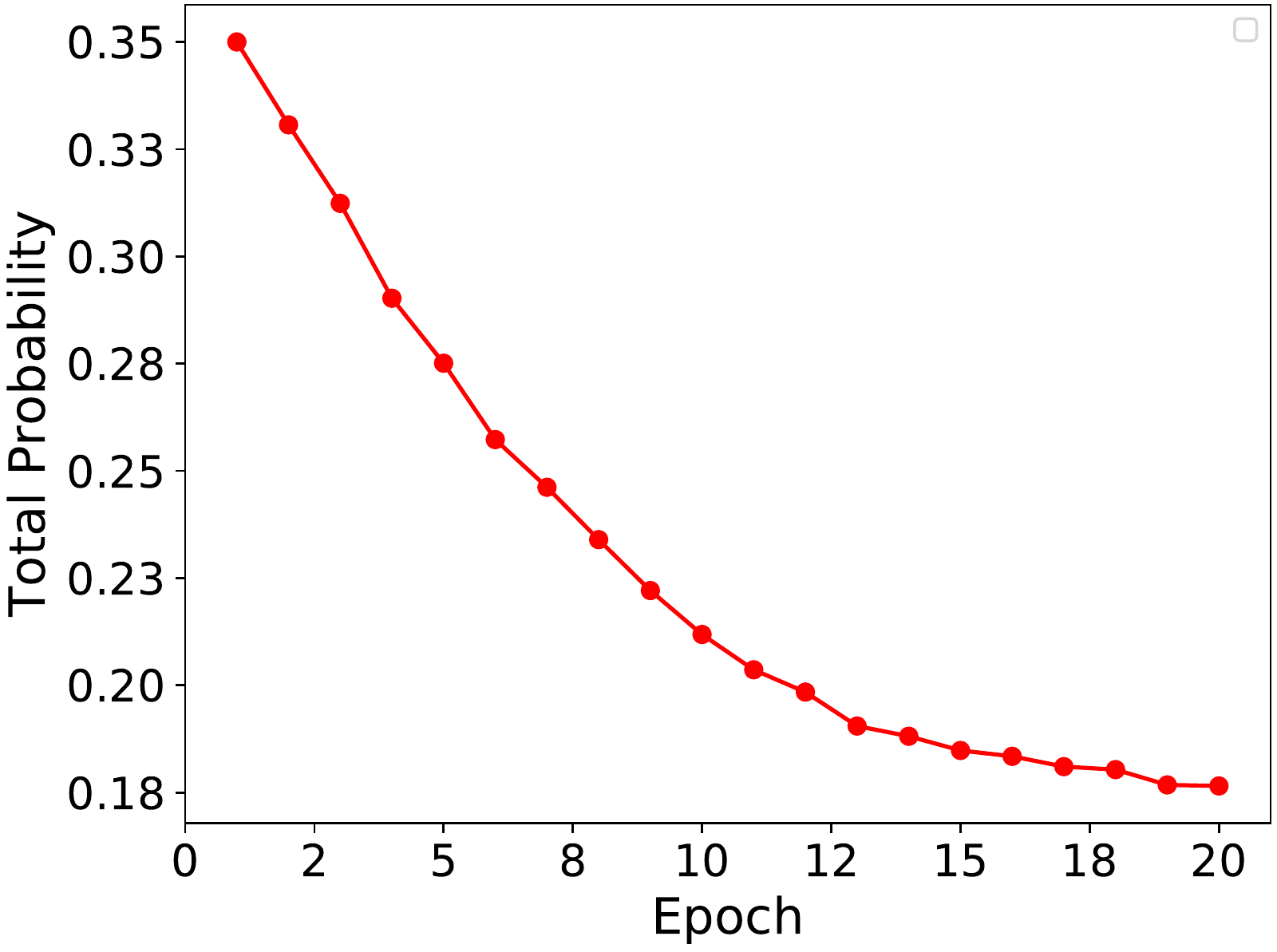}}
	\subfigure[CIFAR-100]{\includegraphics[width=0.48\linewidth, height=0.35\linewidth]{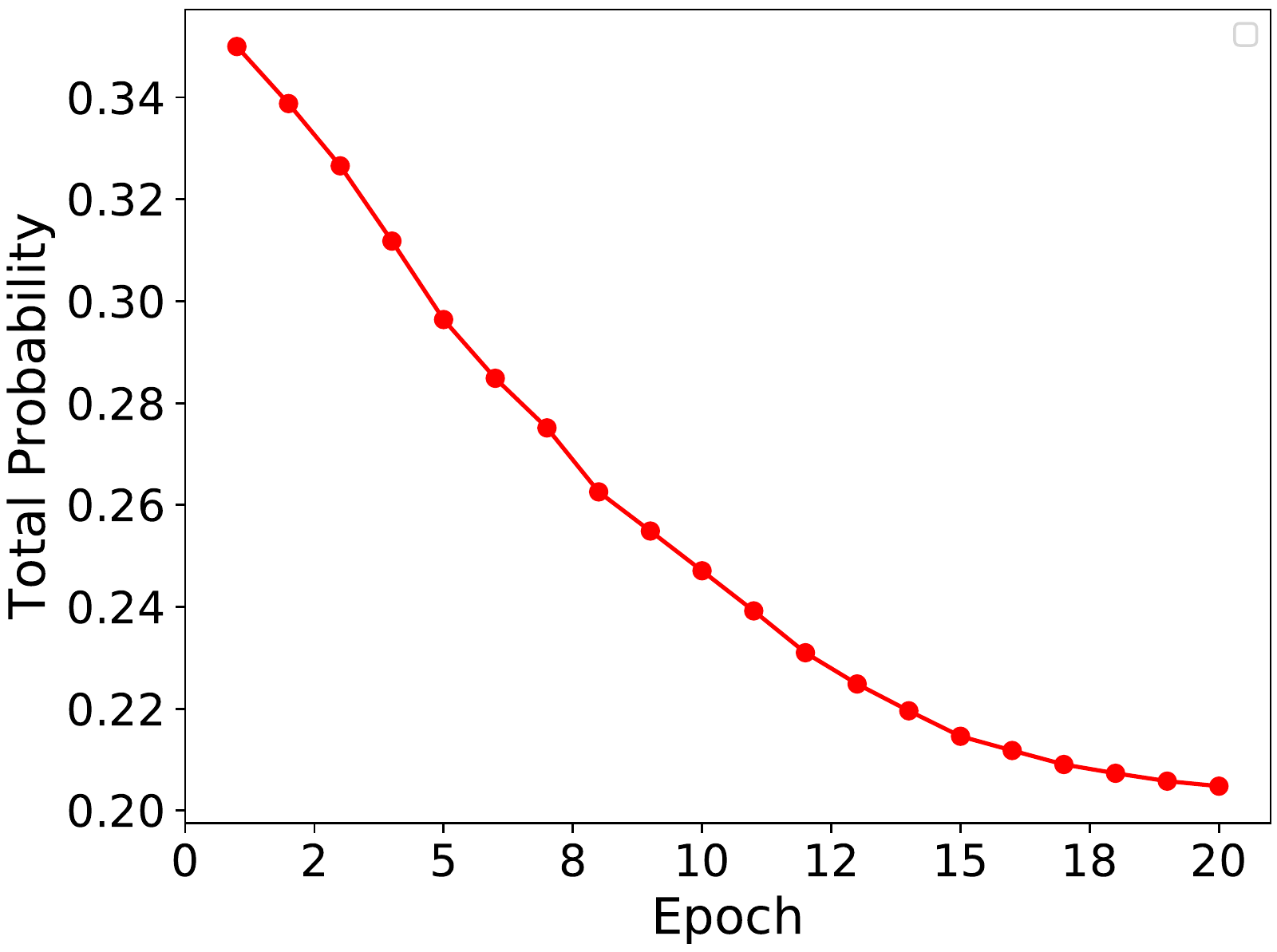}}
	\subfigure[ImageNet]{\includegraphics[width=0.48\linewidth, height=0.35\linewidth]{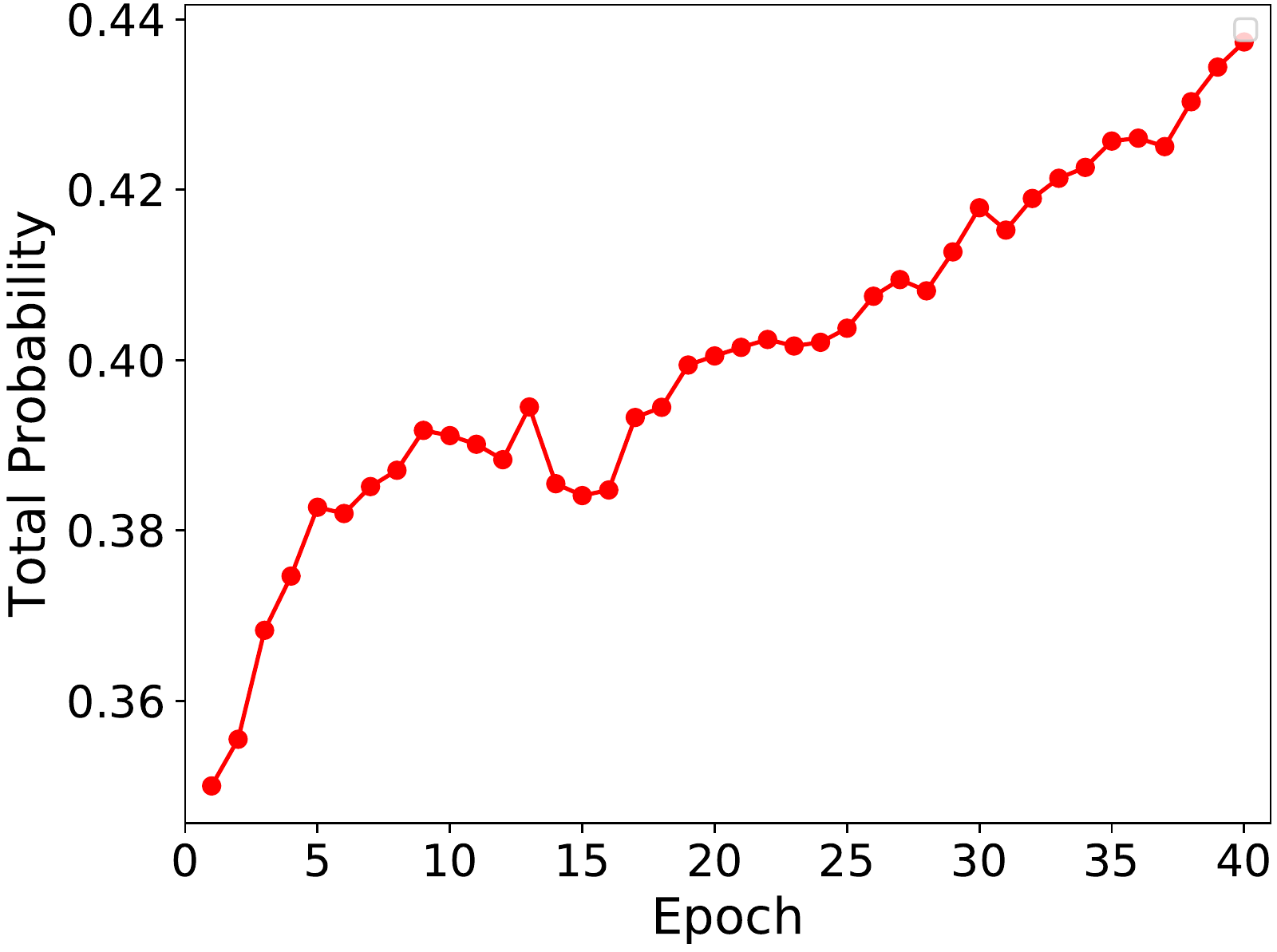}}
	\subfigure[MS-COCO]{\includegraphics[width=0.48\linewidth, height=0.35\linewidth]{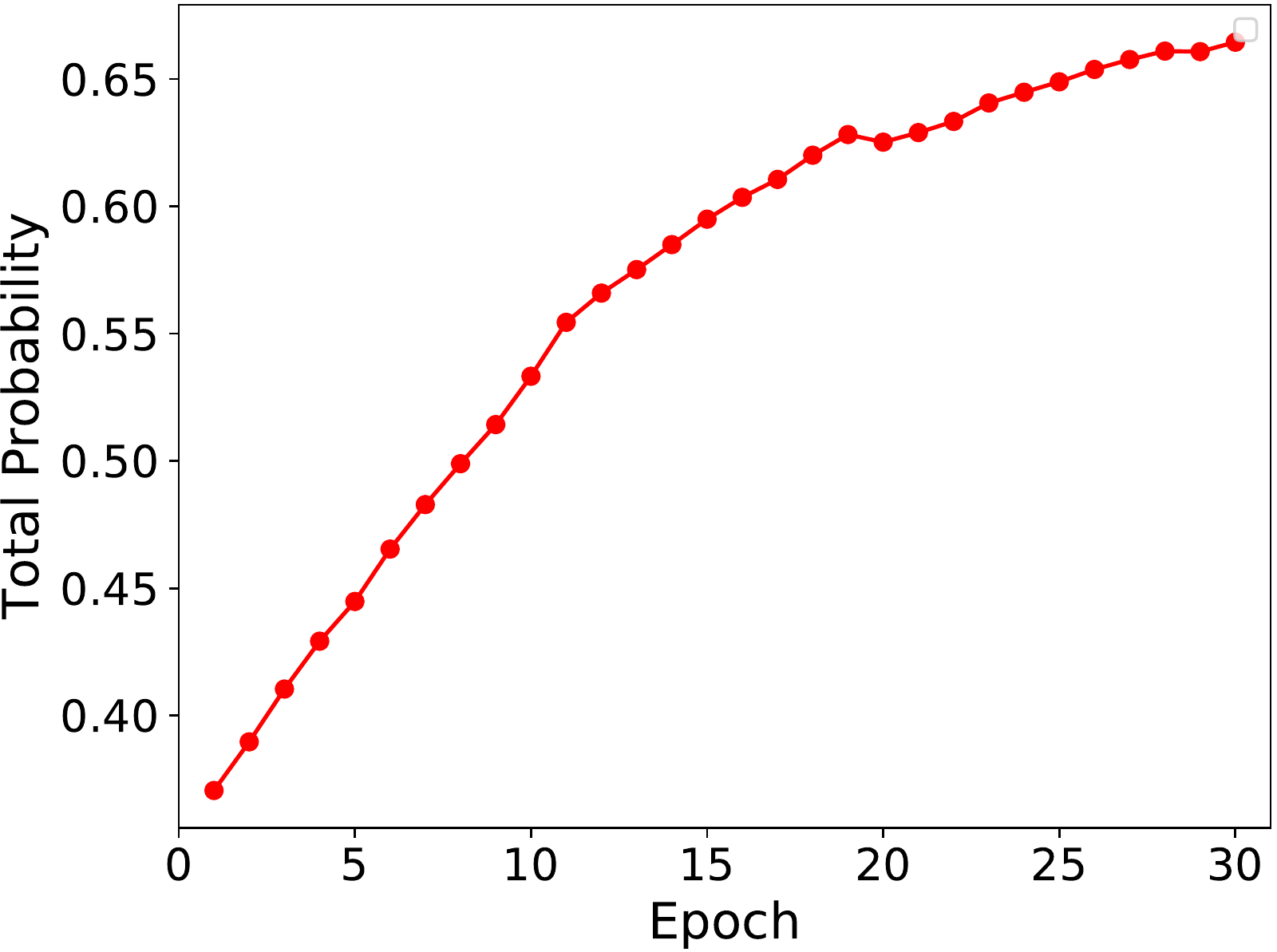}}
	\caption{Total Probability $p_{tp}$ learned on different datasets. The trend of $p_{tp}$ varies among different datasets, which validates the necessity of learning it automatically.}
	\label{fig:ptp_cifar}
	\vspace{-14pt}
\end{figure}

We further visualize the $p_{tp}$ of the augmentations obtained on CIFAR-10 and CIFAR-100 in Fig.\ref{fig:ptp_cifar} (a) \& (b). We can see that the total probabilities keep decreasing, which means our {\ddaa}  prefers less augmentation as epoch number increases on CIFAR-10/100.

\begin{table*}
\small
\caption{
ImageNet test error rates $(\%)$. Every model is trained for 270 epochs if not noted.
}
\label{table:imagenet}
\vspace{3pt}
\centering
\begin{threeparttable}  
\begin{tabular}{l|cccccc|c}
\toprule
Model & Baseline & AA & Fast AA  & Faster AA & DADA & RA$\dag$ & \textbf{\ddaa} \\
\midrule
ResNet-50 & $23.7$ & $22.4$ & $22.4$ & $23.5^{\P}$ & $22.5$ & $22.2$ & $22.3 \pm 0.1$ \\
ResNet-200  & $21.5$ & $20.0$ & $19.4$ & - & - & -& $19.7$\\
\bottomrule
\end{tabular}
\begin{tablenotes}
     \item $\dag$ Reproduced RA with our settings for comparison. $\P$ Trained for 200 epochs.
\end{tablenotes}
\vspace{-12pt}
\end{threeparttable}
\end{table*}
\subsection{ImageNet} 
Due to the different data distributions, the augmentation policy obtained on CIFAR may not be useful when applied to large-scale datasets. For example, Cutout \cite{devries2017improved} shows less improvement on ImageNet than that on CIFAR. Thus we apply our {\ddaa} to directly search on ImageNet.


%


\noindent \textbf{Search Setting}. As for ImageNet, we choose ResNet-18 \cite{he2016deep} as our proxy model for search. The proxy dataset setting is the same as that in AA \cite{cubuk2019autoaugment}. We randomly sample 120 classes from the 1000 classes of ImageNet. 6000 images are sampled (50 images for each class) as the training dataset $S_{train}$ and 1200 images (20 images for each class) are sampled as the validation dataset $S_{val}$.

We run the search for 40 epochs. The training mini-batch size is 32 and validation mini-batch size is 64. At every step, we sample $\hat{L}=2$ augmentation policies according to a uniform distribution. The operation number in a sampled augmentation policy is set to $N_o=2$. The search is carried out on a single RTX 2080Ti GPU. Similar as the CIFAR experiments, we initialize $\p_o$ equally and $p_{tp}$ with 0.35.

\noindent \textbf{Training Setting}. As for training, we train two networks: ResNet-50 and ResNet-200.  Both of these two networks are trained for 270 epochs. Other training and search details are listed in the appendix.
%


%

\noindent \textbf{Result}. The search costs are shown in Table~\ref{table:time} and the test error rates on ImageNet are reported in Table~\ref{table:imagenet}. Our {\ddaa} is on the Pareto optimal curve of the search cost v.s. testing error. The results show the consistent superiority across datasets of different scales.

We also visualize the $p_{tp}$ obtained on ImageNet in Fig.\ref{fig:ptp_cifar} (c). In contrast to CIFAR-10/100 experiment, we can see that $p_{tp}$ obtained on ImageNet increases as training epoch increases. We hypothesize the reason may be the distribution variation between training and validation set of ImageNet data is larger than that in CIFAR, thus aggressive augmentations are needed to fill the distribution gap.

\subsection{Object Detection on COCO} 
In addition to image classification, we further evaluate our {\ddaa} on object detection task. Augmentation search for object detection is rarely tackled. As far as we know, only AA and RA are applied on object detection. \cite{zoph2019learning} specialized AA to object detection by proposing new search space. RA also tried augmentation search within the same search space. The central issue of both methods is they are very time-consuming. AA takes 19200 TPU hours to search for the augmentation policies. As far as we know, our {\ddaa} is the first work to achieve efficient augmentation search for object detection. We implement our search algorithm based on MMDetection \cite{chen2019mmdetection}.

\noindent \textbf{Search Setting}. We use RetinaNet \cite{lin2017focal} with ResNet-50 as backbone to search for augmentation policies. We randomly sample 3000 images as the training dataset $S_{train}$ and 1500 images as the validation dataset $S_{val}$ from \emph{train} split. We run the search for 30 epochs. Both the training mini-batch size and validation mini-batch size are 4 images per GPU. At every step, we will sample $\hat{L}=3$ augmentation policies according to a uniform distribution. The operation number $N_o$ in an sampled augmentation policy is set to 2. As for the augmentation operations, we follow the setting of \cite{zoph2019learning}. The search is carried out on 4 RTX 2080Ti GPUs.

\noindent \textbf{Training Setting}. After obtaining the augmentation policies, we then train two networks following \cite{zoph2019learning} from scratch: RetinaNet with ResNet-50 and ResNet-101 as backbones on the full COCO dataset. The two networks are trained for 150 epochs. 
For fair comparison, we reproduce AA based on our implementation with the searched policy described in \cite{zoph2019learning}. As for RA \cite{cubuk2020randaugment}, we simply copy the result of ResNet-101 from the paper, which is trained for 300 epochs, since we can not reproduce a decent performance.

%

 %
\noindent \textbf{Result}. The search costs and mean Average Precision (mAP) on COCO \emph{val} split are reported in Table~\ref{tab:coco_results}. Our {\ddaa} significantly reduces the search cost for object detection task on COCO, which is $1000 \times$ faster than AA \cite{zoph2019learning} while achieving slightly lower results.
We also visualize the $p_{tp}$ searched on COCO in Fig.\ref{fig:ptp_cifar} (d), and find that $p_{tp}$ increases as training epoch increases, which is similar as the search on ImageNet.
\begin{table}[t]
\caption{Mean Average Precision (mAP) and search cost on COCO dataset for object detection.}
\label{tab:coco_results}  
\centering
\begin{threeparttable} 
\small
\begin{tabular}{llcc}
  \hline
  Backbone & Method &  mAP  & Search Time (hours)\\  
  \hline 
  &Baseline  & 36.9 & 0\\
  ResNet-50 & AA$^\dag$  & 38.4& $19800^{\P}$\\
  &\ddaa\ & $38.1 \pm 0.1$ & $20^*$ \\
  & DADA & not converge$^\star$\\ 
  \hline 
  &Baseline  & 38.6 & 0\\
  ResNet-101 & AA$^\dag$  & 40.0& $19800^{\P}$ \\
  &RA \cite{cubuk2020randaugment}& 40.1& $4600^*$ \\
  &\ddaa\ & 39.8 &  $20^*$ \\
  \hline
\end{tabular}
\begin{tablenotes}
\raggedright
\item $\P$ Evaluated on TPU. * Evaluated on 2080Ti. 
\item$\dag$ Reproduced with searched policy from \cite{zoph2019learning}.
\item$\star$ Confirmed with authors of DADA
\end{tablenotes}
\vspace{-14pt}
\end{threeparttable}
\end{table}

\section{Ablation}
In this section, we further analyze the results and do ablation experiments to show some interesting insights. The experiments in this section are all conducted with Wide-ResNet-28-10 on CIFAR-100 and ResNet-50 on ImageNet.

\noindent
\textbf{The effectiveness of $p_{tp}$.}
A notable difference between {\ddaa} and previous works is that we model the probability of applying augmentation separately.
To prove the superiority of learning an adaptive $p_{tp}$, we manipulate $p_{tp}$ to a fixed value in both search and training to test the results. The results are reported in Table~\ref{table:ptp ablation}, which show that our {\ddaa} could automatically learn the best $p_{tp}$. The results are slightly better than the best fixed $p_{tp}$. Moreover, as illustrated in Fig.\ref{fig:ptp_cifar}, different datasets prefer different $p_{tp}$s. Our method can save the effort of manually tuning $p_{tp}$.

\begin{table}[h]
\centering
\vspace{-6pt}
\caption{Test error rates $(\%)$ v.s. $p_{tp}$.}
\label{table:ptp ablation}
\small
\begin{tabular}{c|ccccc | c }
    \hline
    {$p_{tp}$} & $0.2$ & $0.4$ & $0.6$ &$0.8$& $1.0$ & $p_{tp}^{*}$ \\ 
    \hline
    CIFAR-100 & $16.9$ & $17.0$ & $16.6$ & $18.5$ & $24.7$ & $16.6$\\
    \hline
    ImageNet & $22.8$ & $22.2$ & $22.7$ & $22.7$ & $23.1$ & $22.3$\\
    \hline
\end{tabular}
\vspace{-8pt}
\end{table}

\noindent
\textbf{The effectiveness of dynamic policy.} 
Then we show the necessity of dynamic policies. We fix $p_{tp}$ and $\p_{o}$ to their last searched values, which is similar as AutoAugment. The results are summarized in Table~\ref{table:dyn ablation}. The results show that the dynamic policy indeed improves the performance slightly.

\begin{table}[h]
\centering
\caption{Test error rates $(\%)$ v.s. Dynamic augmentation policy.}
\label{table:dyn ablation}
\small
\begin{tabular}{c|ccc | c }
    \hline
    {Fixed} & $p_{tp}$ & $p_{o}$ & $p_{tp}$, $p_{o}$ & Dynamic \\ 
    \hline
    CIFAR-100 & $16.8$ & $16.7$ & $16.9$  & $16.6$\\
    \hline
    ImageNet & $22.4$ & $22.2$ & $22.4$  & $22.3$\\
    \hline
\end{tabular}
\vspace{-12pt}
\end{table}


\section{Conclusion}

In this paper, we propose {\ddaaFull} ({\ddaa}) for automatic augmentation policy search. {\ddaa} firstly reorganizes the search space with a two-level hierarchy, and then make the search differentibale with one-step gradient update meta-learning and continuous relaxation of the expected training loss. Our {\ddaa} achieves state-of-the-art accuracy and efficiency tradeoff on image classification task on various datasets. Moreover, we are the first work to make efficient search on object detection while achieving competitive performance. 
We believe this concise and effective framework can serve as baseline for many subsequent explorations. 

{\small
\bibliographystyle{ieee_fullname}
\bibliography{egbib}

\begin{thebibliography}{10}\itemsep=-1pt

\bibitem{berthelot2019remixmatch}
David Berthelot, Nicholas Carlini, Ekin~D Cubuk, Alex Kurakin, Kihyuk Sohn, Han
  Zhang, and Colin Raffel.
\newblock \text{ReMixMatch}: Semi-supervised learning with distribution
  alignment and augmentation anchoring.
\newblock {\em arXiv preprint arXiv:1911.09785}, 2019.

\bibitem{berthelot2019mixmatch}
David Berthelot, Nicholas Carlini, Ian Goodfellow, Nicolas Papernot, Avital
  Oliver, and Colin~A Raffel.
\newblock \text{MixMatch}: A holistic approach to semi-supervised learning.
\newblock In {\em NeurIPS}, 2019.

\bibitem{bi2020stabilizing}
Kaifeng Bi, Changping Hu, Lingxi Xie, Xin Chen, Longhui Wei, and Qi Tian.
\newblock Stabilizing {DARTS} with amended gradient estimation on architectural
  parameters.
\newblock {\em ArXiv}, abs/1910.11831, 2019.

\bibitem{chen2020hypernetwork}
Chih-Yang Chen, Che-Han Chang, and Edward~Y Chang.
\newblock Hypernetwork-based augmentation.
\newblock {\em arXiv preprint arXiv:2006.06320}, 2020.

\bibitem{chen2019mmdetection}
Kai Chen, Jiaqi Wang, Jiangmiao Pang, Yuhang Cao, Yu Xiong, Xiaoxiao Li,
  Shuyang Sun, Wansen Feng, Ziwei Liu, Jiarui Xu, Zheng Zhang, Dazhi Cheng,
  Chenchen Zhu, Tianheng Cheng, Qijie Zhao, Buyu Li, Xin Lu, Rui Zhu, Yue Wu,
  Jifeng Dai, Jingdong Wang, Jianping Shi, Wanli Ouyang, Chen~Change Loy, and
  Dahua Lin.
\newblock {MMDetection}: Open mmlab detection toolbox and benchmark.
\newblock {\em arXiv preprint arXiv:1906.07155}, 2019.

\bibitem{chen2020simple}
Ting Chen, Simon Kornblith, Mohammad Norouzi, and Geoffrey Hinton.
\newblock A simple framework for contrastive learning of visual
  representations.
\newblock In {\em ICML}, 2020.

\bibitem{cheng2020improving}
Shuyang Cheng, Zhaoqi Leng, Ekin~Dogus Cubuk, Barret Zoph, Chunyan Bai, Jiquan
  Ngiam, Yang Song, Benjamin Caine, Vijay Vasudevan, Congcong Li, et~al.
\newblock Improving 3d object detection through progressive population based
  augmentation.
\newblock In {\em ECCV}, 2020.

\bibitem{cubuk2019autoaugment}
Ekin~D Cubuk, Barret Zoph, Dandelion Mane, Vijay Vasudevan, and Quoc~V Le.
\newblock \text{AutoAugment}: Learning augmentation strategies from data.
\newblock In {\em CVPR}, 2019.

\bibitem{cubuk2020randaugment}
Ekin~D Cubuk, Barret Zoph, Jonathon Shlens, and Quoc~V Le.
\newblock \text{RandAugment}: Practical automated data augmentation with a
  reduced search space.
\newblock In {\em NeurIPS}, 2020.

\bibitem{deng2009imagenet}
Jia Deng, Wei Dong, Richard Socher, Li-Jia Li, Kai Li, and Li Fei-Fei.
\newblock \text{ImageNet}: A large-scale hierarchical image database.
\newblock In {\em CVPR}, 2009.

\bibitem{devries2017improved}
Terrance DeVries and Graham~W Taylor.
\newblock Improved regularization of convolutional neural networks with cutout.
\newblock {\em arXiv preprint arXiv:1708.04552}, 2017.

\bibitem{gastaldi2017shake}
Xavier Gastaldi.
\newblock Shake-shake regularization.
\newblock {\em arXiv preprint arXiv:1705.07485}, 2017.

\bibitem{grathwohl2017backpropagation}
Will Grathwohl, Dami Choi, Yuhuai Wu, Geoff Roeder, and David Duvenaud.
\newblock Backpropagation through the void: Optimizing control variates for
  black-box gradient estimation.
\newblock In {\em ICLR}, 2018.

\bibitem{hataya2019faster}
Ryuichiro Hataya, Jan Zdenek, Kazuki Yoshizoe, and Hideki Nakayama.
\newblock Faster \text{AutoAugment}: Learning augmentation strategies using
  backpropagation.
\newblock In {\em ECCV}, 2020.

\bibitem{hataya2020meta}
Ryuichiro Hataya, Jan Zdenek, Kazuki Yoshizoe, and Hideki Nakayama.
\newblock Meta approach to data augmentation optimization.
\newblock {\em arXiv preprint arXiv:2006.07965}, 2020.

\bibitem{he2016deep}
Kaiming He, Xiangyu Zhang, Shaoqing Ren, and Jian Sun.
\newblock Deep residual learning for image recognition.
\newblock In {\em CVPR}, 2016.

\bibitem{he2019data}
Zhuoxun He, Lingxi Xie, Xin Chen, Ya Zhang, Yanfeng Wang, and Qi Tian.
\newblock Data augmentation revisited: Rethinking the distribution gap between
  clean and augmented data.
\newblock {\em arXiv preprint arXiv:1909.09148}, 2019.

\bibitem{ho2019population}
Daniel Ho, Eric Liang, Xi Chen, Ion Stoica, and Pieter Abbeel.
\newblock Population based augmentation: Efficient learning of augmentation
  policy schedules.
\newblock In {\em ICML}, 2019.

\bibitem{inoue2018data}
Hiroshi Inoue.
\newblock Data augmentation by pairing samples for images classification.
\newblock {\em arXiv preprint arXiv:1801.02929}, 2018.

\bibitem{jang2016categorical}
Eric Jang, Shixiang Gu, and Ben Poole.
\newblock Categorical reparameterization with \text{Gumbel-Softmax}.
\newblock In {\em ICLR}, 2016.

\bibitem{kostrikov2020image}
Ilya Kostrikov, Denis Yarats, and Rob Fergus.
\newblock Image augmentation is all you need: Regularizing deep reinforcement
  learning from pixels.
\newblock {\em arXiv preprint arXiv:2004.13649}, 2020.

\bibitem{krizhevsky2009learning}
Alex Krizhevsky, Geoffrey Hinton, et~al.
\newblock Learning multiple layers of features from tiny images.
\newblock {\em Technical report}, 2009.

\bibitem{li2020pointaugment}
Ruihui Li, Xianzhi Li, Pheng-Ann Heng, and Chi-Wing Fu.
\newblock \text{PointAugment}: an \text{Auto-Augmentation} framework for point
  cloud classification.
\newblock In {\em CVPR}, 2020.

\bibitem{li2020dada}
Yonggang Li, Guosheng Hu, Yongtao Wang, Timothy~M. Hospedales, Neil~Martin
  Robertson, and Yongxin Yang.
\newblock {DADA:} differentiable automatic data augmentation.
\newblock In {\em ECCV}, 2020.

\bibitem{lim2019fast}
Sungbin Lim, Ildoo Kim, Taesup Kim, Chiheon Kim, and Sungwoong Kim.
\newblock Fast \text{AutoAugment}.
\newblock In {\em NeurIPS}, 2019.

\bibitem{lin2019online}
Chen Lin, Minghao Guo, Chuming Li, Xin Yuan, Wei Wu, Junjie Yan, Dahua Lin, and
  Wanli Ouyang.
\newblock Online hyper-parameter learning for auto-augmentation strategy.
\newblock In {\em ICCV}, 2019.

\bibitem{lin2017focal}
Tsung-Yi Lin, Priya Goyal, Ross Girshick, Kaiming He, and Piotr Doll{\'a}r.
\newblock Focal loss for dense object detection.
\newblock In {\em ICCV}, 2017.

\bibitem{lin2014microsoft}
Tsung-Yi Lin, Michael Maire, Serge Belongie, James Hays, Pietro Perona, Deva
  Ramanan, Piotr Doll{\'a}r, and C~Lawrence Zitnick.
\newblock Microsoft \text{COCO}: Common objects in context.
\newblock In {\em ECCV}, 2014.

\bibitem{liu2018darts}
Hanxiao Liu, Karen Simonyan, and Yiming Yang.
\newblock {DARTS}: Differentiable architecture search.
\newblock In {\em ICLR}, 2019.

\bibitem{mounsaveng2020learning}
Saypraseuth Mounsaveng, Issam Laradji, Ismail~Ben Ayed, David Vazquez, and
  Marco Pedersoli.
\newblock Learning data augmentation with online bilevel optimization for image
  classification.
\newblock {\em arXiv preprint arXiv:2006.14699}, 2020.

\bibitem{netzer2011reading}
Yuval Netzer, Tao Wang, Adam Coates, Alessandro Bissacco, Bo Wu, and Andrew~Y.
  Ng.
\newblock Reading digits in natural images with unsupervised feature learning.
\newblock In {\em NIPSW}, 2011.

\bibitem{ren2018learning}
Mengye Ren, Wenyuan Zeng, Bin Yang, and Raquel Urtasun.
\newblock Learning to reweight examples for robust deep learning.
\newblock In {\em ICML}, 2018.

\bibitem{tang2020onlineaugment}
Zhiqiang Tang, Yunhe Gao, Leonid Karlinsky, Prasanna Sattigeri, Rogerio Feris,
  and Dimitris Metaxas.
\newblock \text{OnlineAugment}: Online data augmentation with less domain
  knowledge.
\newblock In {\em ECCV}, 2020.

\bibitem{tian2020improving}
Keyu Tian, Chen Lin, Ming Sun, Luping Zhou, Junjie Yan, and Wanli Ouyang.
\newblock Improving \text{Auto-Augment} via augmentation-wise weight sharing.
\newblock In {\em NeurIPS}, 2020.

\bibitem{wu2020generalization}
Sen Wu, Hongyang~R Zhang, Gregory Valiant, and Christopher R{\'e}.
\newblock On the generalization effects of linear transformations in data
  augmentation.
\newblock {\em arXiv preprint arXiv:2005.00695}, 2020.

\bibitem{Xie_2020_CVPR}
Cihang Xie, Mingxing Tan, Boqing Gong, Jiang Wang, Alan~L. Yuille, and Quoc~V.
  Le.
\newblock Adversarial examples improve image recognition.
\newblock In {\em CVPR}, 2020.

\bibitem{Xie2019UnsupervisedDA}
Qizhe Xie, Zihang Dai, E. Hovy, Minh-Thang Luong, and Quoc~V. Le.
\newblock Unsupervised data augmentation for consistency training.
\newblock {\em arXiv: Learning}, 2019.

\bibitem{xie2018snas}
Sirui Xie, Hehui Zheng, Chunxiao Liu, and Liang Lin.
\newblock \text{SNAS}: stochastic neural architecture search.
\newblock In {\em ICLR}, 2018.

\bibitem{yun2019cutmix}
Sangdoo Yun, Dongyoon Han, Seong~Joon Oh, Sanghyuk Chun, Junsuk Choe, and
  Youngjoon Yoo.
\newblock \text{CutMix}: Regularization strategy to train strong classifiers
  with localizable features.
\newblock In {\em ICCV}, 2019.

\bibitem{zagoruyko2016wide}
Sergey Zagoruyko and Nikos Komodakis.
\newblock Wide residual networks.
\newblock In {\em BMVC}, 2016.

\bibitem{zhang2018mixup}
Hongyi Zhang, Moustapha Cisse, Yann~N. Dauphin, and David Lopez-Paz.
\newblock Mixup: Beyond empirical risk minimization.
\newblock In {\em ICLR}, 2018.

\bibitem{zhang2020adversarial}
Xinyu Zhang, Qiang Wang, Jian Zhang, and Zhao Zhong.
\newblock Adversarial \text{AutoAugment}.
\newblock In {\em ICLR}, 2020.

\bibitem{zhao2020differentiable}
Shengyu Zhao, Zhijian Liu, Ji Lin, Jun-Yan Zhu, and Song Han.
\newblock Differentiable augmentation for data-efficient gan training.
\newblock In {\em NeurIPS}, 2020.

\bibitem{zhou2020metaaugment}
Fengwei Zhou, Jiawei Li, Chuanlong Xie, Fei Chen, Lanqing Hong, Rui Sun, and
  Zhenguo Li.
\newblock Metaaugment: Sample-aware data augmentation policy learning, 2021.

\bibitem{zhou2020data}
Sharon Zhou, Jiequan Zhang, Hang Jiang, Torbj{\"o}rn Lundh, and Andrew~Y Ng.
\newblock Data augmentation with mobius transformations.
\newblock {\em arXiv preprint arXiv:2002.02917}, 2020.

\bibitem{zoph2019learning}
Barret Zoph, Ekin~D Cubuk, Golnaz Ghiasi, Tsung-Yi Lin, Jonathon Shlens, and
  Quoc~V Le.
\newblock Learning data augmentation strategies for object detection.
\newblock In {\em ECCV}, 2020.

\bibitem{zoph2016neural}
Barret Zoph and Quoc~V Le.
\newblock Neural architecture search with reinforcement learning.
\newblock In {\em ICLR}, 2017.

\end{thebibliography}
}
\appendix
\clearpage
\section{Addtional Experiments}
\subsection{Experiment on SVHN}
\noindent \textbf{Search Setting}. We also search augmentation on SVHN \cite{netzer2011reading}. Similar as the experiments on CIFAR-10/100, we first search augmentation on a proxy task with a small network, Wide-ResNet-40-2 on part of the dataset for 20 epochs. We split 3000 images for training dataset $S_{train}$ and 3000 images for $S_{val}$. The training mini-batch size is 32 while the validation mini-batch size is 256. At each step, we sample $\hat{L}=3$ augmentation policies randomly according to a uniform distribution. The number of operation $N_o$ in an augmentation policy is 2. The search is carried out on a single RTX 2080Ti GPU. As for initialization, we initialize $\p_o$ equally and $p_{tp}$ as 0.35. 

$\alpha_{tp}$ and $\alo$ are updated with Adam optimizers. The learning rate for $\alo$ is 0.005, and that for $\alpha_{tp}$ is 0.001. For the 2 optimizers, we set $\beta_1 = 0.5,\ \beta_2 = 0.999$. The network parameters of Wide-ResNet-40-2 are updated with SGD optimizer with momentum as 0.9. The learning rate is 0.05, with cosine decay, and the weight decay is 0.0005.

\noindent \textbf{Training Setting}. After search, we apply the searched augmentation to Wide-ResNet-28-10. We train the network for 160 epochs on the full training set and report the performance evaluated on test set.  We set initial learning as 0.005, batch size as 128, momentum as 0.9, weight decay as 0.001, and cosine learning rate decay. 

\noindent \textbf{Result}. The search cost and test error rate are shown in Table.\ref{table:svhn}. We run the experiment for three times and report the average test error rate. Compared with other efficient automatic augmentation methods, our \ddaa~ can also achieve comparable performance and efficiency.
\begin{table*}
	\small
	\caption{
		SVHN test error rates $(\%)$ and search cost (GPU hour). WRN is shorthand of Wide-ResNet.
	}
	\vspace{3pt}
	\label{table:svhn}
	\centering
	\begin{tabular}{l|cccccccc|c}
		\toprule
		& Baseline & Cutout & AA & PBA & Fast AA & RA & Faster AA & DADA & \textbf{\ddaa} \\
		\midrule
		
		\textbf{Error} & & & & & & & & & \\
		WRN-28-10 & $1.5$ & $1.3$ & $1.1$ & $1.2$ & $1.1$ & $1.0$ & $1.2$ & $1.2$ & $1.2$ \\
		\midrule
		\textbf{Cost} & & & & & & & & & \\
		WRN-28-10 & - & - & $1000$ & $1$ & $1.5$ & - & $0.06$ & $0.1$ & $0.1$ \\
		\bottomrule
	\end{tabular}
\end{table*}

\subsection{The effectiveness of policy number $\hat{L}$.}

Our DDAS achieves efficient search because it can search for good augmentation policies with limited sampled augmentation policies at each step ($\hat{L} = 2,\ 3$). We further explore the effectiveness of sampled augmentation policy number $\hat{L}$ on the performance of searched policies. We search for augmentation policies with different $\hat{L}$ values and show the results in Table~\ref{table:number ablation}. We can see that simply increasing the sampled augmentation policy number $\hat{L}$ does not increase the performance.

\begin{table}[h]
	\centering
	\caption{Sampled augmentation policy number $\hat{L}$ ablation.}
	\label{table:number ablation}
	\small
	\begin{tabular}{c|ccc}
		\hline
		$\hat{L}$ & $3$ & $7$ & $11$ \\ 
		\hline
		Error & $16.6$ & $16.8$ & $17.1$\\
		\hline
	\end{tabular}
	\vspace{-8pt}
\end{table}

\section{Implemntation Details}
\subsection{Image Classsification}
\noindent \textbf{Operations}. Here we list the augmentation operations used for image classification.
\begin{table}[H]
	\footnotesize
	\centering
	\def\arraystretch{1.0}
	\begin{tabular}{lll}
		$\bullet\;$ \texttt{Identity} & $\bullet\;$ \texttt{AutoContrast} &
		$\bullet\;$ \texttt{Equalize} \\ $\bullet\;$ \texttt{Rotate} &
		$\bullet\;$ \texttt{Solarize} & $\bullet\;$ \texttt{Color} \\
		$\bullet\;$ \texttt{Posterize} &
		$\bullet\;$ \texttt{Contrast} & $\bullet\;$ \texttt{Brightness} \\
		$\bullet\;$ \texttt{Sharpness} & $\bullet\;$ \texttt{Shear-x/y} & $\bullet\;$ \texttt{Smooth}\\
		$\bullet\;$ \texttt{Translate-x/y} & $\bullet\;$ \texttt{Invert} & $\bullet\;$ \texttt{Blur} \\
		$\bullet\;$ \texttt{FlipLR}  & $\bullet\;$ \texttt{FlipUD} 
	\end{tabular}
\end{table}

Similar as AA \cite{cubuk2019autoaugment}, the operations are from PIL, a popular Python image library.\footnote{https://pillow.readthedocs.io/en/5.1.x/} In addition, we add \texttt{Cutout} to search space of ImageNet, and use it defaultly with region size as 16 pixels on CIFAR-10/100 and SVHN. 

\noindent \textbf{Magnitude}. As for the magnitude, we follow RA \cite{cubuk2020randaugment} and use the same linear scale for indicating the magnitude (strength) of each transformation. As mentioned in Method, we discretely sampled magnitude value, and the selected magnitude values for CIFAR-10/100, SVHN and ImageNet are listed in Table \ref{tb:mags}.


\subsection{Object Detection}

For object detection, we use the operations in \cite{zoph2019learning}. The magnitude setting keeps the same as \cite{zoph2019learning}. The selected magnitude values for COCO is listed in Table \ref{tb:mags}.

\begin{table}[h]
	\centering
	\caption{Manitudes sampled for different datasets}
	
	\newcommand{\tabincell}[2]{\begin{tabular}{@{}#1@{}}#2\end{tabular}}  
	\label{tb:mags}
	\small
	\begin{tabular}{c | c }
		\hline
		{Dataset} & $mag$ \\ 
		\hline
		\hline
		CIFAR-10/ CIFAR-100/ SVHN                                 & $\{2,6,10,14\}$ \\
		\hline
		ImageNet  &   $\{7,14\}$ \\
		\hline
		COCO  &  $\{4,6,8\}$\\
		\hline
	\end{tabular}
	
\end{table}

\section{Experiments Details}

\subsection{CIFAR-10/100 \& Sanity Check}

\noindent \textbf{Search Setting}. As for search both $\alpha_{tp}$ and $\alo$ are updated with Adam optimizers. The learning rate for $\alo$ is 0.005, and that for $\alpha_{tp}$ is 0.001. For both of the 2 optimizers, we set $\beta_1 = 0.5,\ \beta_2 = 0.999$. The network parameters of Wide-ResNet-40-2 are updated with SGD optimizer with momentum as 0.9. The learning rate is 0.05, with cosine decay, and the weight decay is 0.0005.

As for Sanity Check, the settings are basically similar as that of CIFAR-10/100 experiments. The difference is that we only use $N_o = 1$, magnitude as $2$ and $p_{tp}$ initialized as 0.75.

\noindent \textbf{Training Setting}. Both of the 2 training networks are trained with SGD optimizer whose momentum is 0.9. The batch size is 128 and the cosine LR schedule is adopted for all the models. For Wide-ResNet-28-10, we set initial learning as 0.1 and weight decay as 0.0005. For Shake-Shake(26 2x96d), we set initial learning as 0.01 and weight decay as 0.001.  The $p_{t}(1),...,p_{t}(T_{max})$ are smoothed with a mean filter whose size $F_s=2$.

\subsection{ImageNet}

\noindent \textbf{Search Setting}. Both $\alpha_{tp}$ and $\alo$ are optimized with Adam optimizer. The learning rate for $\alo$ is 0.003, and that for $\alpha_{tp}$ is 0.002. For the 2 optimizers, we set $\beta_1 = 0.5,\ \beta_2 = 0.999$. The network parameters of ResNet-18 are updated with SGD optimizer. The learning rate is 0.01, with cosine decay, the momentum is 0.9 and the weight decay is 0.0001. 

\noindent \textbf{Training Setting}. As for training, the 2 networks are trained with SGD optimizer whose momentum is 0.9. We set initial learning rate as 0.2, batch size as 512, momentum as 0.9, weight decay as 0.0001 and step learning decay by 0.1 at epoch 90, 180 and 240. For fair comparison we reproduce RA on ResNet-50 with our training setting and report the result. The $p_{t}$s are smoothed with a mean filter whose size $F_s=6$.

\subsection{COCO}
\noindent \textbf{Search Setting}. Both $\alo$ and $\alpha_{tp}$ are updated with Adam optimizers. The learning rate for $\alo$ is 0.001, and that for $\alpha_{tp}$ is 0.001. And we set $\beta_1 = 0.5,\ \beta_2 = 0.999$ for the 2 optimizers. The parameters of RetinaNet are updated with SGD optimizer. The learning rate is 0.04, with step learning decay at epoch 24 and 28, the momentum is 0.9 and the weight decay is 0.0001.

\noindent \textbf{Training Setting}. Both of the 2 training networks are trained with SGD optimizer whose momentum is 0.9. We set the initial learning rate as 0.08, batch size as 64, momentum as 0.9, weight decay as 0.0001 and step learning rate decay at epoch 120 and 140.

\end{document}